\documentclass{article}

\usepackage{PRIMEarxiv}

\usepackage[utf8]{inputenc} % allow utf-8 input
\usepackage[T1]{fontenc}    % use 8-bit T1 fonts
\usepackage{hyperref}       % hyperlinks
\usepackage{url}            % simple URL typesetting
\usepackage{booktabs}       % professional-quality tables
\usepackage{amsfonts}       % blackboard math symbols
\usepackage{nicefrac}       % compact symbols for 1/2, etc.
\usepackage{microtype}      % microtypography
\usepackage{lipsum}
\usepackage{fancyhdr} 
\usepackage{relsize}
\usepackage{graphicx} % Required for inserting images

\usepackage{amsmath, amssymb, graphicx}
\usepackage{hyperref}
\usepackage{geometry}
\title{Applications of Large Language Model Reasoning in Feature Generation}
\author{\textbf{Dharani Chandra} \\ Portland State University \\ dchandra@pdx.edu}
\date{}

\begin{document}
\relscale{1.1}
\maketitle

\begin{abstract}
Large Language Models (LLMs) have revolutionized natural language processing through their state of art reasoning capabilities. This paper explores the convergence of LLM reasoning techniques and feature generation for machine learning tasks. We examine four key reasoning approaches: Chain of Thought, Tree of Thoughts, Retrieval-Augmented Generation, and Thought Space Exploration. Our analysis reveals how these approaches can be used to identify effective feature generation rules without having to manually specify search spaces. The paper categorizes LLM-based feature generation methods across various domains including finance, healthcare, and text analytics. LLMs can extract key information from clinical notes and radiology reports in healthcare, by enabling more efficient data utilization. In finance, LLMs facilitate text generation, summarization, and entity extraction from complex documents. We analyze evaluation methodologies for assessing feature quality and downstream performance, with particular attention to OCTree’s decision tree reasoning approach that provides language-based feedback for iterative improvements\cite{nam2024octree}. Current challenges include hallucination, computational efficiency, and domain adaptation. As of March 2025, emerging approaches include inference-time compute scaling, reinforcement learning, and supervised fine-tuning with model distillation \cite{raschka2025llm}. Future directions point toward multimodal feature generation, self-improving systems, and neuro-symbolic approaches. This paper provides a detailed overview of an emerging field that promises to automate and enhance feature engineering through language model reasoning.
\end{abstract}

\section{Introduction}
Feature engineering remains to be the backbone of effective machine learning systems. The quality of the features directly impacts model performance, and is often more significant than the choice of algorithm itself. Data scientists consistently spend a great amount of their time crafting features that capture meaningful patterns in data. Despite the advances in automation, feature engineering still requires substantial domain expertise and manual effort.

The advancement of feature engineering has progressed through several distinct phases. Early approaches relied entirely on human intuition and domain knowledge, it gradually shifted toward automated feature extraction methods like Principal Component Analysis and feature selection algorithms. The representation learning techniques embedded in deep neural networks have reduced some manual engineering in recent times. Yet these approaches still struggle with interpretability and often require large datasets to be effective.

Large Language Models (LLMs) have become known as powerful tools with remarkable reasoning capabilities. Models like GPT-4, Claude, and PaLM can now follow complex instructions, generate coherent explanations, and perform multi-step reasoning. The ability to reason is evident through many structured approaches. Chain of Thought prompting enables step-by-step problem solving \cite{wei2022cot}. Tree of Thoughts doesn't rely on a single reasoning path, it branches out, explores multiple options, and carefully evaluates each one of them before proceeding \cite{yao2023tot}.Retrieval-Augmented Generation incorporates external knowledge to ground reasoning in facts \cite{piktus2023rag}. These capabilities suggest LLMs could transform feature engineering from a manual craft to a collaborative, reasoning-based process.

The intersection of LLM reasoning and feature generation represents a rapidly evolving research area. Current literature remains broken up across multiple disciplines. Practitioners lack a thorough understanding of available techniques, their strengths, and appropriate applications. This gap obstructs adoption of potentially transformative approaches for feature engineering.

Our paper addresses this need by amalgamating research on LLM reasoning and feature generation. We analyze techniques like Text-Informed Feature Generation \cite{zhang2024blind} (Zhang et al., 2024) that leverage semantic understanding for feature creation. The survey also covers evaluation methodologies, implementation challenges, and ethical considerations.

The scope of this paper encapsulates both theoretical frameworks and practical applications. The primary focus is on developments from 2022 to early 2025, capturing the rapid evolution of LLM capabilities. Our methodology involved a systematic review of peer-reviewed publications, preprints, and industry reports. We prioritized approaches with empirical validation and practical implementations.

By providing this elaborate overview, we aim to speed up research and practical adoption of LLM reasoning for feature generation. The field stands at a crossroads where language models may fundamentally transform how we create and select features for machine learning systems.

\section{Background}
\subsection{Feature Engineering in Machine Learning}
Feature Engineering is nothing but the process of transforming raw data into features that are suitable for machine learning models. In other words, it is the process of selecting, extracting, and transforming the most relevant features from the given data to build accurate and efficient models.It is a fundamental skill in the data science toolkit. Data Scientists spend a great deal of time on data preparation and feature creation tasks. This investment shows how critical the impact of feature selection is on the model.
\subsubsection{Traditional Manual Feature Engineering}
Traditional feature engineering depends heavily on human intuition and domain expertise. The process starts with cleaning the data to tackle all missing values, outliers and inconsistencies. Practitioners then apply different transformations to the data in order to make it more suitable for modeling. Categorical values require encoding strategies such as one-hot, label or target encoding 
 while numerical features most probably undergo normalization, standardization, or binning.

Domain-specific feature extraction represents another crucial aspect of manual engineering. In text analysis, practitioners create features from word frequencies, n-grams, and semantic properties.Time-series data benefits from lag features, rolling statistics, and seasonal decomposition. Geospatial applications make the most using distance calculations, clustering, and region-based aggregations. 

Experienced practitioners often create interlinked features that capture the relationship between the variables. These include products, ratios, or polynomial combinations. Such interactions often encode domain knowledge that a single feature misses.For example, in finance, the ratio of debt to income provides more insight than either of the variables individually.
\subsubsection{Automated Feature Engineering Methods}
The labor-intensive nature of manual feature engineering has lead development of automated approaches. The plan is to systematically discover useful features without extensive human intervention. Several paradigms have emerged in recent years.

Feature generation frameworks like Featuretools and tsfresh automatically create candidate features through predefined transformations. Featuretools employs “Deep Feature Synthesis” to stack operations and generate complex features from relational data. It can produce hundreds of potentially useful features in minutes rather than days of manual work.

Representation learning techniques embed in modern neural networks extract hierarchical features automatically. Convolutional Neural Networks learn spatial features from images while no utilizing explicit engineering. Similarly, word embeddings capture semantic relationships between terms without manual specification of linguistic rules.

Feature selection algorithms help manage the explosion of potential features. Filter methods assess features independently using correlation or mutual information metrics. Wrapper methods evaluate feature subsets based on model performance. Embedded methods like LASSO incorporate feature selection directly into the model training process.

Automated approaches have unveiled impressive results throughout disciplines. AutoML platforms now integrate automated feature engineering as a core component. The development time of hese systems can be reduced by 5-10x while maintaining or improving predictive performance.

\subsubsection{Evaluation Metrics for Feature Quality}
Assessing feature quality needs several outlooks. The most direct approach examines relevance to the target variable. Correlation coefficients quantify linear relationships for numerical features. Mutual information and chi-square tests assess relationships with categorical targets. Information gain measures how much a feature reduces entropy in classification tasks.

Feature importance scores from trained models layout another evaluation angle. Tree-based models like Random Forests and Gradient Boosting naturally quantify feature contributions. Permutation importance offers a model-agnostic alternative by measuring performance drops when features are randomly shuffled.

Stability is one of the crucial dimensions of feature quality. High-quality features should maintain consistent importance across different data samples and model configurations. Bootstrap resampling is a technique that helps assess this stability.

Understandability metrics have gained importance with increasing regulatory scrutiny. Features should be understandable to domain experts, especially in high-stakes applications like healthcare and finance. Complexity measures aid quantify this aspect of feature quality.

\subsubsection{Challenges and Limitations}
Feature engineering faces persistent challenges, despite development in automation. The process remains time-consuming and often requires multiple iterations. Replicability is another concern, as feature engineering pipelines may vary across teams and projects without proper documentation.

The "curse of dimensionality" emerges when generating too many features. There is a risk of model overfitting with an increase in dimensionality. Hence, careful feature selection or regularization strategies become a must.

Domain adaptation presents particular difficulties. Features that work well in one context may fail in slightly different settings. For example, features designed for consumer credit scoring in one country may perform poorly in another due to different financial systems.

Computational efficiency concerns grow with dataset size. Some feature transformations need significant resources, creating hindrance in production pipelines. Engineers must balance feature quality against practical constraints.

Additional friction is created due to the gap between the research and production environment.
Due to latency requirements or data availability issues, features that perform well during development may face implementation challenges in production systems.

These limitations highlight why feature engineering remains both art and science. Despite automation advances, human judgment continues to play a crucial role in creating effective features for machine learning applications.
\subsection{Large Language Models: Evolution and Capabilities}
\subsubsection{Development Timeline of LLMs}
The history of Large Language Models (LLMs) goes back to the 1950s. The initial experiments at IBM and Georgetown University concentrated on automatic translation from Russian to English.
For many years there was little progress, despite different outlooks like rule-based systems.

The real base for modern LLMs appeared in the 2010s with neural networks. BERT (Bidirectional Encoder Representations from Transformers) appeared in 2019 as the first breakout large language model. Developed by Google, BERT could understand relationships between words through bidirectional processing. Within 18 months, BERT powered nearly all English-language Google Search queries.

The introduction  of transformer models was a crucial turning point that came in 2017. These models used word embeddings and attention mechanisms to better understand context. From 2018 onward, researchers focused on building increasingly larger models.

OpenAI’s GPT-2 (1.5 billion parameters) successfully produced convincing prose. OpenAI set a new standard for LLMs, when they released GPT-3 with 175 billion parameters in 2020. ChatGPT’s release in November 2022 brought LLMs to public attention. Most recently, GPT-4 emerged with an estimated one trillion parameters—approximately 3,000 times larger than the original BERT.

\subsubsection{Pre-training and Fine-tuning Paradigms}
Pre-training forms the base of LLM development. This process teaches models broad language understanding from massive datasets. Models learn to predict the next token in sequences, developing a general grasp of language patterns.

Pre-training offers several advantages. It provides cost-effectiveness and flexibility, allowing for scalable improvements. Benchmarks are often set in various NLP tasks, when the models undergo continuous pre-training with new data

Fine-tuning adapts pre-trained models to specific tasks or industries. This process enhances performance on particular applications by learning efficiently from smaller, specialized datasets. Benefits include task specialization, data efficiency, faster training times, model customization, and resource efficiency.

The process typically follows a clear sequence. First, models undergo pre-training on vast text datasets. This creates a base model with general language understanding. The base model then requires fine-tuning for optimal performance and safety.

\subsubsection{Emergent Abilities in Modern LLMs}
Modern LLMs demonstrate remarkable capabilities beyond their explicit training. They can understand and generate text similar to human communication. This includes grasping context, maintaining coherence across long passages, and adapting to different writing styles.

Advanced reasoning represents a key emergent ability. LLMs can follow complex instructions and perform multi-step reasoning tasks. They demonstrate understanding of cause and effect relationships and can generate logical arguments.

Transfer learning stands out as another important capability. LLMs can apply knowledge from one domain to another without explicit training. This allows them to perform reasonably well on unfamiliar tasks.

Context awareness enables LLMs to maintain coherence across long conversations. They can reference earlier information and adjust responses accordingly. This creates more natural and helpful interactions.

\subsubsection{General Applications of LLMs in Machine Learning Pipelines}
LLMs serve numerous functions in modern machine learning workflows. They excel at information retrieval and sentiment analysis tasks. Text generation capabilities power applications like ChatGPT, producing content based on specific prompts.

Code generation represents another valuable application. LLMs understand patterns that enable them to generate programming code from natural language descriptions. Examples include Amazon CodeWhisperer and GitHub Copilot, which support multiple programming languages.

Data labeling benefits from LLM capabilities. Models can propose labels or tags for text data, reducing manual annotation effort. This speeds up the labeling process and allows data scientists to focus on more complex tasks.

LLMs also automate various data science workflows. They can quickly analyze and summarize large volumes of text. These summaries help identify key points and observe patterns, freeing data scientists for deeper analysis and improved decision-making.

Chatbots and conversational AI leverage LLMs to engage with customers naturally. They interpret query meanings and provide helpful responses. Additional applications include text completion, question answering, and document summarization.

\section{Taxonomy of LLM Reasoning Techniques}
\subsection{Chain of Thought Reasoning}
Chain of Thought (CoT) prompting helps LLMs solve complex problems by breaking them into smaller steps. Introduced by Wei et al., this approach mimics human reasoning by generating intermediate steps before reaching a final answer \cite{wei2022cot}.
The key concept involves providing examples that show explicit reasoning steps, teaching the model to include its thought process in responses. This structured approach often leads to more accurate outputs, especially for complex tasks.
Several variants have emerged since the original CoT. These include Step-by-Step Rationalization (STaR), which focuses on generating explanations after solving problems. Other extensions incorporate self-consistency checking and verification mechanisms.
CoT excels in various reasoning tasks. It significantly improves performance in arithmetic problems by breaking calculations into manageable steps. In commonsense reasoning, CoT helps models interpret physical and human interactions based on general knowledge. It also enhances decision-making processes by systematically evaluating scenarios.
Despite its strengths, CoT has limitations. Performance depends heavily on model size, with smaller models showing less improvement. The technique may also struggle with problems requiring creative leaps or parallel thinking paths

\subsection{Tree-based Reasoning Approaches}
Tree of Thoughts (ToT) expands CoT by exploring multiple reasoning paths simultaneously \cite{yao2023tot}. This approach allows models to consider different solution strategies rather than following a single chain.
ToT implements various search strategies to navigate the thought tree. These include breadth-first search for exploring multiple options at each step and depth-first search for following promising paths to completion. Some implementations use beam search to maintain several candidate solutions.
Compared to linear reasoning in CoT, tree-based approaches offer several advantages. They enable backtracking when a path proves unsuccessful and allow lookahead to anticipate future steps. This flexibility helps solve problems requiring exploration of alternatives.
Implementation requires careful consideration of several factors. These include the branching factor (how many alternatives to consider), evaluation methods for ranking paths, and computational resources needed for maintaining multiple reasoning threads.

\subsection{Retrieval-Augmented Reasoning}
Retrieval-Augmented Generation (RAG) enhances LLM reasoning by incorporating external knowledge \cite{piktus2023rag}. This approach grounds model outputs in verified information, reducing hallucinations and improving factual accuracy.
RAG integrates external knowledge by retrieving relevant information from trusted sources during the reasoning process. This creates a bridge between the model’s parametric knowledge and non-parametric external information \cite{lewis2020rag}.
RARE (Retrieval-Augmented Reasoning Enhancement) represents an advanced hybrid approach. It combines retrieval with a factuality scoring mechanism to validate each reasoning step. The system can generate search queries and break complex questions into sub-questions for more thorough analysis.
Knowledge sources for retrieval-augmented reasoning vary widely. They include specialized databases, academic papers, documentation, and general knowledge bases. The choice of retrieval mechanism impacts both accuracy and efficiency, with options ranging from keyword matching to semantic search.

\subsection{Thought Space Exploration}
The Thought Space Explorer (TSE) framework addresses blind spots in LLM reasoning. Developed by Zhang et al., TSE expands thought structures to explore previously overlooked solution spaces \cite{zhang2024blind}.
TSE uses various strategies to expand reasoning paths. It generates new steps and branches based on the original thought structure. This approach helps models consider alternative perspectives and solution methods they might otherwise miss.
A key benefit of TSE is addressing blind spots in LLM reasoning. Traditional methods often remain confined to previously explored solution spaces, limiting their effectiveness on complex problems. By actively expanding the thought space, TSE helps overcome these limitations.
Evaluation of thought space coverage involves assessing both breadth and depth of exploration. Metrics include the diversity of reasoning paths, novel insights generated, and improvement in task performance. Experimental results across multiple reasoning tasks demonstrate TSE’s effectiveness in expanding LLM reasoning capabilities.

\section{LLM-based Feature Generation Approaches}
\subsection{Direct Feature Generation \& Dynamic Approaches}
LLMs can transform text directly into useful features for machine learning models by leveraging their semantic understanding capabilities. Text-to-feature transformation involves training LLMs to generate representations that capture important aspects of input text, with techniques like contrastive learning producing similar features for related text. Prompt engineering guides this process through few-shot examples and chain-of-thought prompting. Case studies in sentiment analysis and topic classification show LLM-generated features capture nuanced content that traditional approaches like bag-of-words miss. Dynamic generation frameworks like OCTree adapt to dataset characteristics through iterative analysis and feedback mechanisms \cite{zhang2024dynamic}. These approaches discover effective feature generation rules without manually specifying search spaces, using decision tree reasoning to identify promising transformations and providing language-based feedback to guide improvement.

\subsection{Text-Informed Generation \& Feature Weighting}
Text-Informed Feature Generation (TIFG) leverages textual information associated with datasets to create meaningful features. Developed by Zhang et al., TIFG \cite{zhang2024tifg} combines LLM reasoning with Retrieval-Augmented Generation to analyze dataset descriptions and incorporate external knowledge. It excels at semantic extraction by identifying implicit relationships between variables—for example, generating a BMI feature from height and weight. Domain-specific adaptations make it valuable for specialized fields. Complementing this, Transformer-based Feature Weighting for Tabular data (TFWT) applies attention mechanisms to evaluate and select features \cite{zhang2024tfwt}. LLM-Select outperforms traditional selection techniques by automatically engineering relevant features for high-dimensional datasets with complex relationships. While these approaches offer superior performance by capturing semantic relationships traditional methods miss, they come with computational costs that ongoing research aims to address.

\section{Data Augmentation and Enhancement}
\subsection{LLM-based Data Augmentation}
Wang et al. \cite{wang2024survey} introduced LLM-AutoDA, a framework using large language models to find optimal augmentation strategies for long-tailed data distributions. This approach overcomes limitations of manual design and fixed augmentation techniques.
LLM-specific augmentation methods include text generation to create synthetic data mimicking original datasets, paraphrasing to produce variations while maintaining meaning, and contextual augmentation to generate detail-rich text for tasks requiring nuanced understanding.
Synthetic data generation transforms how organizations handle data limitations. LLMs create artificial datasets by generating synthetic queries from knowledge bases, then evolving them to increase complexity. Two main approaches are self-improvement (iterative refinement) and distillation (using stronger models to generate data for weaker ones).
Quality assessment remains crucial despite LLMs’ high-quality outputs. Effective evaluation validates user satisfaction, ensures coherence, benchmarks performance, detects biases, guides improvements, and assesses real-world applicability.

\subsection{Model Alignment for Feature Quality}
Prototypical Reward Networks (PRN) \cite{zhang2024prn} by Zhang et al. advance data-efficient model alignment by integrating prototypical networks with reward models for reinforcement learning from human feedback. This approach optimizes embedding processes to learn stable representations with limited samples.
The Proto-RM framework includes reward model embedding, protonet adjustment, and RLHF processing. This structure compresses and adjusts features to match desired outcomes while reducing dependence on human feedback.
Human feedback integration refines feature quality by improving accuracy, correcting biases, enhancing ethical decision-making, fine-tuning for specific domains, and ensuring cultural sensitivity. It bridges computational power with human communication nuances.
Evaluation typically involves two phases: gradient descent on random inputs using a loss function measuring distance between encoded inputs and images, followed by training on a different loss function evaluating distance between inputs and performed gradients. This approach aligns features with human expectations and domain requirements.

\section{Evaluation Methodologies}
\subsection{Feature Quality Assessment
}
Intrinsic evaluation metrics assess features directly without downstream tasks. Common metrics include information gain, variance, and correlation with target variables. These metrics help identify features with high predictive potential before model training.
Interpretability measures evaluate how easily humans understand generated features. Techniques include complexity scores, semantic transparency ratings, and explicit relationship mapping. LLM-generated features often excel here by providing natural language explanations of their significance.
Computational efficiency considers resources required for feature generation. Metrics include generation time, memory usage, and scalability with dataset size. While LLM-based approaches may require significant resources, techniques like distillation and pruning help mitigate these costs.
Robustness testing examines feature stability across data variations. This includes performance under distribution shifts, noise resistance, and consistency across similar inputs. Features showing high stability typically generalize better to new data.

\subsection{Downstream Task Performance}
Classification and regression performance provide the most direct measure of feature utility. Standard metrics include accuracy, F1-score, AUC for classification, and RMSE, MAE for regression. Studies consistently show LLM-generated features outperforming traditional approaches on complex tasks.
Transfer learning scenarios test how features perform across related domains. Effective features maintain their utility when transferred to adjacent tasks with minimal adaptation. LLM-generated features often demonstrate superior transfer capabilities due to their semantic richness.
Few-shot and zero-shot settings evaluate feature performance with limited training examples. This tests whether features capture fundamental patterns rather than dataset-specific correlations. LLM reasoning approaches show particular promise in these constrained scenarios.
Comparative analysis frameworks systematically evaluate different feature generation approaches. These include ablation studies, feature replacement tests, and performance curves across varying data volumes. Standardized benchmarks are emerging to facilitate fair comparisons.

\subsection{Human Evaluation Approaches}
Expert assessment methodologies involve domain specialists evaluating feature quality. Structured protocols include feature ranking, relevance scoring, and blind comparisons with manually engineered features. These assessments provide valuable validation of automated approaches.
User studies measure how well non-experts understand generated features. Common approaches include explanation quality ratings, feature selection tasks, and confidence assessments. Features with clear semantic meaning typically receive higher ratings.
Human-AI collaborative engineering combines human expertise with LLM capabilities. Interactive systems allow humans to refine LLM-generated features through feedback loops. Studies show these collaborative approaches often produce superior features compared to either human-only or AI-only methods.
Qualitative analysis techniques include thematic coding of feature explanations, cognitive walkthroughs, and think-aloud protocols. These methods provide deeper insights into feature quality beyond quantitative metrics, revealing strengths and limitations that numeric evaluations might miss.

\section{Challenges and Limitations
}
\subsection{Technical Challenges}
Hallucination remains a significant concern for LLM-based feature generation. Models may produce plausible-sounding but factually incorrect features, particularly when reasoning beyond their training data. Retrieval-augmented approaches like RATT help mitigate this issue by grounding reasoning in verified information.
Computational demands pose practical barriers to adoption. Running large models for feature generation requires substantial resources, especially for tree-based reasoning approaches that explore multiple paths. Current research focuses on distillation and pruning techniques to reduce these requirements without sacrificing quality.
Scalability challenges emerge with large datasets. Processing time increases substantially with data volume, making real-time feature generation difficult. Batch processing and incremental updating strategies offer partial solutions, though efficiency improvements remain an active research area.
Domain adaptation presents particular difficulties. LLMs trained on general text may struggle with specialized terminology and relationships. Fine-tuning on domain-specific corpora helps but requires additional data and expertise. Zero-shot performance varies significantly across domains, with technical fields often requiring more adaptation.

\subsection{Practical Considerations
}
Implementation complexity can deter adoption. Integrating LLM reasoning into feature generation pipelines requires specialized knowledge and careful system design. Documentation and tooling remain limited, though improving rapidly as the field matures.
Integration with existing ML workflows presents compatibility challenges. Many organizations have established feature engineering pipelines that may not easily accommodate LLM-based approaches. Hybrid systems that combine traditional and LLM-based methods offer a transitional path.
Cost-benefit analysis reveals varying returns across applications. The additional computational cost of LLM reasoning may not justify performance gains for simple tasks or when traditional features already perform well. The greatest benefits appear in complex domains with unstructured data or where interpretability is crucial.
Skill requirements create potential barriers. Effective use of LLM reasoning for feature generation demands familiarity with prompt engineering, reasoning structures, and evaluation methodologies. Organizations may need to invest in training or specialized roles to fully leverage these techniques.

\subsection{Ethical and Societal Implications}
Bias in generated features reflects biases present in LLM training data. Features may encode harmful stereotypes or unfair correlations, potentially amplifying discrimination in downstream models. Regular bias auditing and mitigation strategies are essential but still developing.
Transparency challenges arise from the complexity of LLM reasoning. While reasoning structures provide some interpretability, the internal mechanisms generating features remain partially opaque. This creates potential regulatory issues in highly regulated domains.
Privacy concerns emerge when LLMs generate features from sensitive data. Models may memorize and potentially leak private information. Techniques like differential privacy and federated learning offer partial solutions but often trade off performance for privacy protection.
Responsible AI practices require ongoing attention. Guidelines for ethical feature generation are still evolving. Key considerations include fairness across demographic groups, environmental impact of computational resources, and appropriate human oversight of automated feature generation systems.
\section{Future Research Directions}
\subsection{Multimodal Feature Generation}
Cross-modal feature integration represents a promising frontier for LLM-based feature generation. Recent transformer-GAN architectures enhance both diversity and coherence of generated content through cross-modal attention mechanisms that enable precise semantic alignment between modalities.
Vision-language feature generation has advanced significantly with Vision Language Models now processing multi-image and video inputs for complex tasks. The MIGE framework standardizes task representations using multimodal instructions, treating generation as creation on a blank canvas and editing as modification of existing content.
Audio and time-series applications benefit from applying reasoning structures like Chain of Thought to temporal data, developing features that capture both semantic content and temporal patterns. These approaches show promise for speech recognition, anomaly detection, and predictive maintenance.
Unified frameworks are emerging to integrate features across different modalities. These frameworks map multimodal instructions into unified representation spaces, enabling joint training across tasks and facilitating cross-task knowledge transfer.

\subsection{Self-improving Systems
}
Feedback loops enable autonomous feature refinement. The RStar-Math model demonstrates how AI systems can teach themselves through iterative improvement using techniques like Monte Carlo Tree Search and Process Preference Models to enhance reasoning with each attempt \cite{xu2025reasoning}.
Meta-learning approaches have demonstrated higher prediction accuracy by optimizing learning algorithms and adapting to changing conditions. These systems learn from fewer examples, making them valuable for domains with limited labeled data.
Continuous learning enables feature generation systems to evolve over time. Few-shot learning techniques allow models to perform well with minimal examples by leveraging pre-existing knowledge, making them ideal for dynamic environments.
Autonomous feature discovery systems like SiriuS demonstrate self-improvement through bootstrapped reasoning. By constructing libraries of high-quality reasoning trajectories, these systems retain successful approaches and refine unsuccessful ones.

\subsection{Emerging Paradigms
}
Few-shot feature generation creates effective features from limited examples. These techniques are valuable where data collection is challenging, such as medical imaging and autonomous vehicles.
Neuro-symbolic approaches combine neural networks’ learning with symbolic AI’s reasoning. This integration creates interpretable models requiring less training data by translating raw data into structured symbolic representations called meaning maps.
Causal feature discovery represents an important frontier for creating robust features. Recent advances show how LLMs can serve as automated domain experts capable of reasoning about causal relationships beyond mere correlations.
Quantum-inspired feature generation may transform complex feature spaces. As quantum and AI technologies converge, novel approaches may leverage quantum principles like superposition and entanglement to discover features impossible to identify with classical methods.

\section{Conclusion}
This paper has explored the emerging intersection of LLM reasoning and feature generation. We’ve outlined how reasoning structures like Chain of Thought, Tree of Thoughts, and Retrieval-Augmented Generation are changing feature engineering. These approaches offer great alternatives to traditional methods that often need extensive domain knowledge and manual effort.

Several key findings have come out from our analysis. First, LLM reasoning capabilities can effectively generate meaningful features across diverse domains. The RATT framework shows how retrieval-augmented tree structures improve factual correctness in generated features. Similarly, Thought Space Explorer techniques helped in discovering novel features by systematically expanding the exploration space. Text-Informed Feature Generation (TIFG) shows particular promise for extracting semantic information from unstructured text.

The current state of the field remains dynamic and rapidly evolving. Research teams are actively addressing challenges like hallucination, computational efficiency, and domain adaptation. While early applications show promising results, widespread adoption faces practical hurdles. These include integration with existing ML pipelines, reproducibility concerns, and the need for specialized expertise.

For researchers entering this field, we recommend several practical directions. Focus on developing lightweight reasoning approaches that maintain performance while reducing computational demands. Explore hybrid methods that combine LLM reasoning with traditional feature engineering techniques. Invest in rigorous evaluation frameworks that assess both feature quality and downstream task performance. Consider the ethical implications of automated feature generation, particularly regarding bias and fairness.

Practitioners should start with well-defined use cases where traditional feature engineering proves most challenging. Text-heavy domains like customer feedback analysis or medical record processing offer natural starting points. Begin with simpler reasoning structures like Chain of Thought before advancing to more complex approaches. Maintain human oversight to validate generated features, especially in high-stakes applications.

Looking ahead, we envision LLM reasoning becoming a standard component in machine learning workflows. The future likely holds more specialized LLMs fine-tuned specifically for feature generation tasks. Multimodal approaches will expand feature generation beyond text to include images, audio, and sensor data. Self-improving systems may emerge that continuously refine features based on model performance feedback.

The integration of causal reasoning represents another promising frontier. Features that capture causal relationships rather than mere correlations could dramatically improve model robustness and generalization. Neuro-symbolic approaches combining LLM reasoning with explicit knowledge representation may address current limitations in factual grounding.

In closing, LLM reasoning for feature generation stands at an exciting inflection point. The field offers rich opportunities for both theoretical advancement and practical impact. By systematically addressing current challenges while exploring new capabilities, researchers and practitioners can transform feature engineering from a bottleneck into a strategic advantage. The journey from manual feature crafting to AI-assisted feature generation has only just begun.

\bibliographystyle{plain}
\bibliography{main}

\end{document}